\pdfoutput=1

\documentclass[11pt]{article}

\usepackage[final]{EMNLP2022}
\usepackage{amsmath}
\usepackage{amssymb}
\usepackage{graphicx}
 \usepackage{enumitem} 
\usepackage{times}
\usepackage{latexsym}

\usepackage{caption}
\usepackage{subcaption}

\usepackage[T1]{fontenc}

\usepackage{microtype}
\usepackage{comment}
\usepackage{tabularx}
\usepackage{dblfloatfix}
\usepackage{float}

\usepackage[utf8]{inputenc}

\usepackage{microtype}

\usepackage{inconsolata}

\newcommand\blfootnote[1]{%
  \begingroup
  \renewcommand\thefootnote{}\footnote{#1}%
  \addtocounter{footnote}{-1}%
  \endgroup
}

\title{Named Entity Recognition in Industrial Tables using \\ Tabular Language Models}

\author{Aneta Koleva\textsuperscript{1,2}{*}, Martin Ringsquandl\textsuperscript{1}{*}, Mark Buckley\textsuperscript{1}, Rakebul Hasan \textsuperscript{1} \and Volker Tresp\textsuperscript{1,2} \\
         \textsuperscript{1}Siemens AG, \textsuperscript{2}Ludwig-Maximilians University \\ first\_name.last\_name@siemens.com}

\begin{document}
\maketitle
\blfootnote{$^\ast$ Equal Contribution.}

\begin{abstract}
Specialized transformer-based models for encoding tabular data have gained interest in academia. Although tabular data is omnipresent in industry, applications of table transformers are still missing.
In this paper, we study how these models can be applied to an industrial Named Entity Recognition (NER) problem where the entities are mentioned in tabular-structured spreadsheets.
The highly technical nature of spreadsheets as well as the lack of labeled data present major challenges for fine-tuning transformer-based models.
Therefore, we develop a dedicated table data augmentation strategy based on available domain-specific knowledge graphs.

We show that this boosts performance in our low-resource scenario considerably. 
Further, we investigate the benefits of tabular structure as inductive bias compared to tables as linearized sequences.
Our experiments confirm that a table transformer outperforms other baselines and that its tabular inductive bias is vital for convergence of transformer-based models.

\end{abstract}

\section{Introduction}

There has been growing interest in developing special model designs intended to capture tabular structure \cite{turl, TaBERT, tapas, tuta}. A recent survey named these models tabular language models (TaLMs) and provided an overview of the different architectures and pretraining objectives \cite{Dong2022}. One of the downstream tasks where TaLMs are applicable is table interpretation (TI) with its sub-tasks: entity linking, column type annotation and relation extraction \cite{turl}. Most TaLMs for TI use BERT as the backbone language model (LM) for encoding the content of table cells and aggregate their representations on different levels (cell, row or column) depending on the task. 

\begin{figure*}[t]
    \centering
    \includegraphics[width=\textwidth]{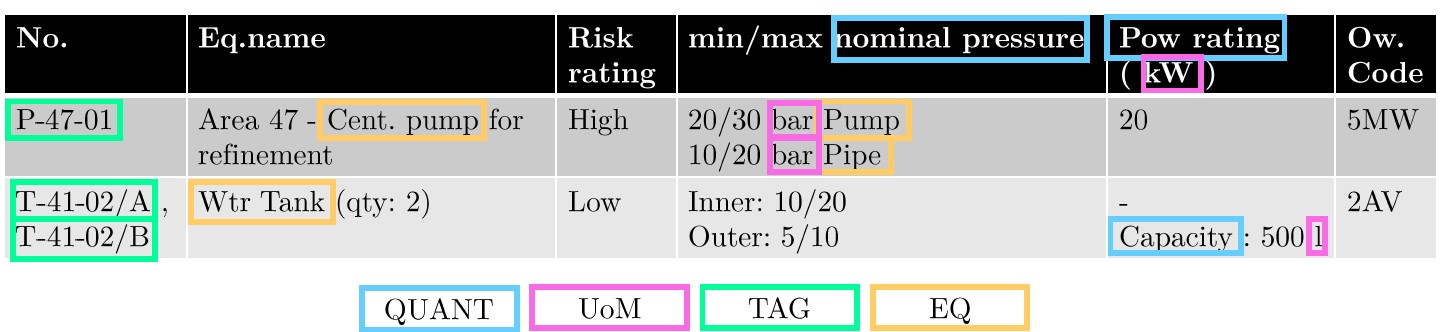}
    \caption{Example table from an industrial plant equipment spreadsheet. Boxes represent NER annotations.}
    \label{fig:example_table}
\end{figure*}

Although tabular data is omnipresent in industry, TaLMs such as table transformers, have not found their way into industrial applications yet. One reason being the nature of data stored in industrial tables which is different and more dynamic than data in academic datasets where the schema of the table is consistent and each cell contains a single entity of one type \cite{cutrona_vincenzo_2020_4246370}. As shown in Figure~\ref{fig:example_table}, industrial tables contain multiple \textit{sub-cell} entities from different types, hence the TaLMs which provide cell-level aggregation are not sufficient. In this direction, we formulate the problem of sub-cell named entity recognition (NER) in tables using TaLMs. 

Another challenge is that tabular data in industry is often lacking labels, especially labels reflecting the high variance across examples. Due to the very technical and domain-specific nature only experts can effectively provide such labels, which is -- for most tasks -- too expensive. These low-resource scenarios are challenging for statistical NLP models and usually prohibit fine-tuning of large-scale transformer-based models.   
A popular strategy to remedy low-resource scenarios is data augmentation (DA) \cite{Simard1996}, which allows to increase data diversity without having to collect new examples.
Common DA techniques in NLP range from using external knowledge such as WordNet \cite{Zhang2015}, machine-translation models for back-translation \cite{dataAugment2} or mixing of examples inspired from computer vision \cite{Yun2019}.
An empirical study by \cite{Longpre2020} showed that applying off-the-shelf DA techniques \cite{Sennrich2016, Wei2019} for fine-tuning of LM like BERT or RoBERTa bring little to no improvement and become even less beneficial in cross-domain settings \cite{tapas, Zhong2020}. 
These studies emphasize the challenge of developing domain-specific DA techniques which would help improve the existing pretrained~transformer~models.  

Although, there are no domain-specific DA techniques applicable to a tabular dataset,
in many industrial domains there exist external resources which can be exploited for creating augmented tables. In this paper we study a DA technique for industrial spreadsheet tables leveraging publicly available resource based on an industrial standard.
Specifically, the contributions of this paper are the following:
\begin{itemize}
    \item We introduce a table transformer model for sub-cell NER, \textsc{TabNER}, and provide an industrial use case as a motivation for this. To the best of our knowledge, this is the first attempt to solve NER in tables with TaLMs.
    \item We develop a novel DA technique for semantically consistent augmentation of tables based on domain-specific knowledge graphs. 
    \item We empirically show that the inductive bias of TaLMs is valuable and combined with our DA technique boosts the performance by 9\% compared to a sequential model.
\end{itemize}

\section{Industry NER Use Case}
As motivation for tabular NER in an industrial context, we describe a real-world dataset from which information about industrial plant equipment, such as actuators, sensors, vessels, etc. and their physical quantities should be extracted. This information is typically collected and maintained by engineers in spreadsheets. The spreadsheets are roughly organized in a tabular format, as can be seen from the example table in Figure~\ref{fig:example_table}. In these spreadsheets, each row typically represents information about one or multiple equipment instances. Some columns represent relevant physical properties of these equipments, while others are non-informative. However, the engineers do neither comply to a fixed schema nor to unified spelling of equipment or properties. 
The goal is to automatically extract relevant entities for creating a structured specifications of the plant equipment. We phrase this problem as NER task with the following types of entities. The type \textit{TAG} refers to a systematic identifier of an equipment. There are some conventions for generating equipment tags (e.g. NORSOK, KKS), but most plant operators customize them and some sheets do not contain identifiers at all. 
Type \textit{EQ} is for surface names of equipment types. The type \textit{QUANT} refers to the physical properties/quantities describing the functional specifications of equipment and the type \textit{UoM} stands for unit of measurement.

\begin{table}
 
\centering
\begin{tabularx}{\linewidth}{l|c *{5}r}
\textbf{Dataset} & $\mu_{\text{tok}}$ & $\sigma_{\text{tok}}$ & $K_{\text{tok}}$ & $\mu_{\text{col}}$ & $\sigma_{\text{col}}$ &  \\ \hline 
SemTab & 2  & 2.5 & 132.2 & 4.5 & 1.9  \\ 
Plant & 2.6 & 3.7 & 585.3  & 16.3  & 21.6  \\
\end{tabularx}
\caption{Dataset statistics: academic vs. industry.}
\label{tab:tab_stats}
\end{table}

\begin{figure*}[t]
    \centering
    \includegraphics[width=0.95\textwidth]{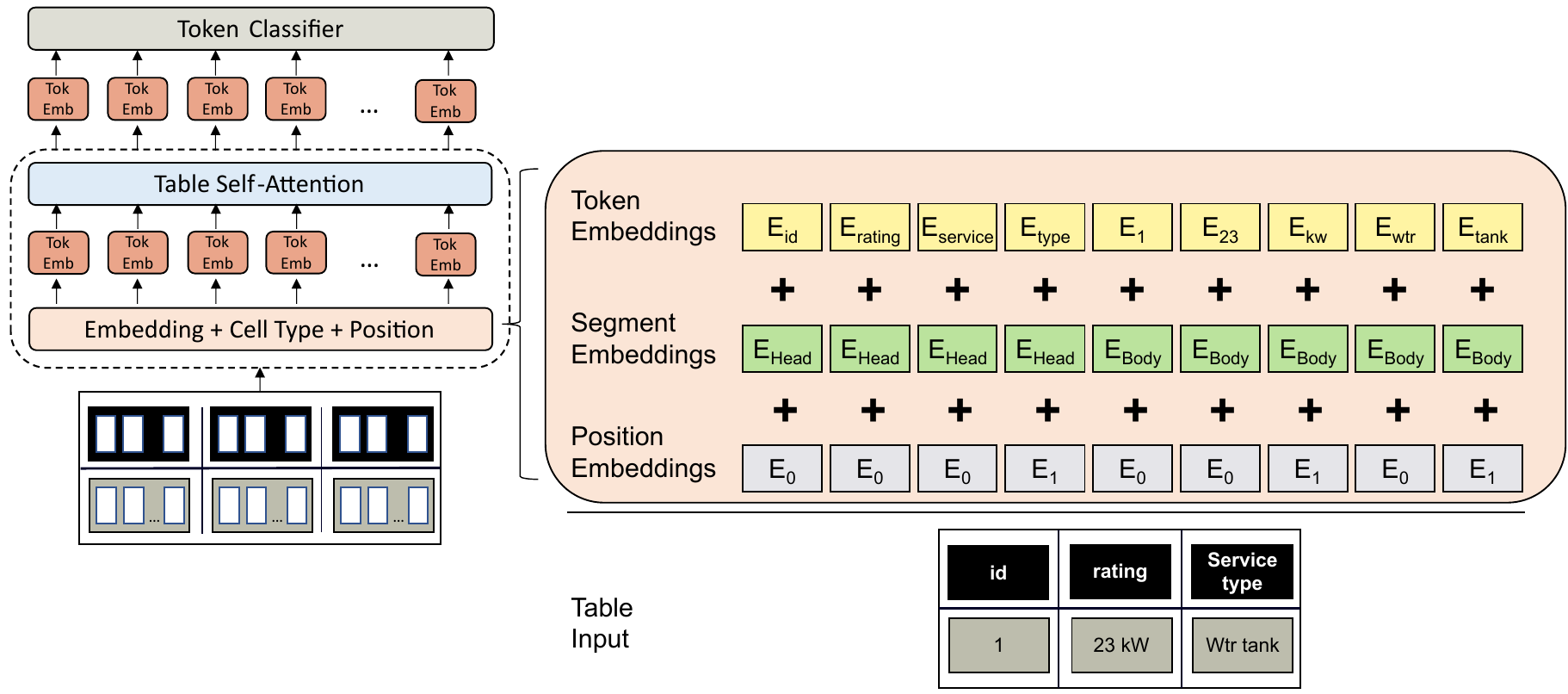}
    \caption{Input modifications to vanilla transformer to encode tokens with tabular structure.}
    \label{fig:bert_input}
\end{figure*}

\paragraph{Table Statistics}
It is not obvious why performing NER in tables would benefit from sophisticated language models. In fact, looking at common tabular benchmark datasets, such as the ones used in the SemTab challenge \cite{cutrona_vincenzo_2020_4246370}, detecting entities is usually very straightforward. Since all tokens in a cell are assumed to represent a single entity, sub-cell NER is an unnecessary step and we only need to perform entity/cell linking. 
Looking at the example table in Figure~\ref{fig:example_table}, however, gives the impression that these industrial spreadsheets are very differently structured from common benchmarks. There can be quite some text and even multiple sub-cell entities in a single cell.
Table~\ref{tab:tab_stats} supports this impression with statistical evidence. The average number of tokens per cell, $\mu_{\text{tok}}$, is 30\% higher in our industrial dataset compared to a dataset from SemTab. Further, its standard deviation $\sigma_{\text{tok}}$ and the Kurtosis $K_{\text{tok}}$, show that there is more variance due to the much longer tail of the distribution of number of tokens in the plant tables. Even more obvious is the difference at the column level where the tables in the SemTab challenge contain on average 4 times less columns ($\mu{\text{col}}$) than the tables describing plant equipment specification with much lower variance as well. This suggests that every token in our NER task has a much broader intra- and inter-cell context.

\section{Related Work}

There has been some research focused on extracting entities and their quantities from web tables. Ibrahim et al. \cite{Ibrahim2016} phrased this problem as entity linking using a table-biased Markov random field and distant supervision. 

Wu et al. \cite{Wu2018} employed BiLSTM models to encode rich-format documents (unstructured text, headings, tables) that mention electronic components, quantities and units of measure. They used hand-crafted labeling functions for collecting (weakly) labeled entities and relations which can be used as weak supervision. 

A recent work on table classification \cite{Koleva2021} compared TaLMs like TaBERT \cite{TaBERT} versus non-contextual word embedding methods for generating table vector representations. TURL \cite{turl} uses a Transformer \cite{Vaswani2017} with table-specific attention mechanism which has been pre-trainined and fine-tuned towards solving the tasks of table interpretation: column type annotation, entity linking and relation extraction. However, this methods generates representations on a cell level and therefore can not be applied for solving our NER problem. 

We are not aware of any work that uses TaLMs for sub-cell table NER in an industrial setting. 

\paragraph{Data Augmentation}
Recently, many different DA techniques have been proposed with the purpose to solve low-resource issues in NLP by generating new examples from existing datasets. For a comprehensive overview on the different DA techniques, we refer the readers to the recent survey by Feng et al. \cite{surveyDA}. 

Several simple and effective DA techniques for NER are presented by \cite{dataAugment1}. However, these techniques are not directly applicable to the industrial tabular data since they rely on domain-agnostic linguistic resources like WordNet. Similarly, methods for sequence labeling, such as backtranslation \cite{dataAugment2} can not be applied to tabular data because the content of the tables are mostly facts and not full sentences.

\section{Method}
We now define the table NER problem and outline how we encode tokens in tables using TaLMs.  

We define a~table~as~a~tuple $T=(C,H)$, where $C = \{c_{1,1}, c_{1,2},  \dots c_{i,j} , \dots , c_{n,m} \}$ is the set of table body cells for $n$ rows and $m$ columns.~Every cell $c_{i,j} = \left( w_{c_{i,j},1}, w_{c_{i,j},2}, \dots, w_{c_{i,j},t} \right)$ is a sequence~of tokens of length $t$.~The table header $H = \{h_1, h_2, \dots, h_m\}$ is the set of corresponding $m$ column header cells, where $h_j =~\left(w_{{h_j},1}, w_{{h_j},2}, \dots , w_{{h_j},q}\right)$ is a sequence~of header tokens with length $q$.~We use $T_{[i,:]}$ to refer to the $i$-th row ($H = T_{[0,:]}$) and $T_{[:,j]} = \{h_j, c_{1,j}, \dots , c_{n,j} \}$ to refer to the $j$-th~column~of~$T$. 

Each labeled cell has an NER-tag sequence: $(y_1, y_2, \dots, y_{|cell|})$, where each $y_i \in \mathcal{Y}$. We use IO tags, thus $\mathcal{Y}$ is $\{\textit{O}\} \cup \{\textit{I-ENT}\}$, where \\ $\textit{ENT} \in \{\textit{TAG},\textit{EQ},\textit{QUANT},\textit{UoM}\}$.

\subsection{\textsc{TabNER} Model}
Compared to the existing TaLMs such as TaBERT \cite{TaBERT}, TURL or TAPAS \cite{tapas} which generate cell-level representations, we propose a simple modification to the vanilla transformer \cite{Vaswani2017} which allows us to use almost any pre-trained transformer\footnote{\url{huggingface.co} token classification models that take a custom 2D \texttt{attention\_mask}} to obtain a (sub-cell) token-level representations for a table.  

Our \textsc{TabNER} model consists of a token encoder layer {\scshape Enc} and a classification layer. 
A conceptual architecture of the table token input encoding is shown in Figure~\ref{fig:bert_input}, where token vector representations for each token in the linearized table are generated by aggregating the token embeddings, the segment embeddings, and position embeddings. The segment indicates if a token is part of the head or the body (instead of the 1\textsuperscript{st} / 2\textsuperscript{nd} sentence semantics) and the position encoding is done on a cell-level, so it restarts from 0 for every cell in body $C$ and header $H$:
\begin{eqnarray*}
    \textit{pos}(T) =& (\textit{pos}(h_i), \dots, \textit{pos}(c_{i,j})) \\
    \textit{pos}(cell) =& (0, \dots, |cell|)
\end{eqnarray*}

Similarly as in TURL, we use a table attention mask (visibility matrix) $\alpha_{i,j}$, but on token-level instead of cell-level. This mask allows every token to attend exclusively to tokens which are either in the same row or in the same column. $\alpha_{i,j} $ is a symmetric binary matrix defined as: 
\[
    \alpha_{i,j} = 
    \begin{cases}
    1 & \text{if } \textit{col}(i) = \textit{col}(j) \lor \textit{row}(i) = \textit{row}(j), \\
    0              & \text{otherwise},
    \end{cases}
\] where $\textit{row } (\textit{col})$ are functions that map linearized token indices back to row (column) indices in the table.

The output of the token encoder layer is a sequence of token representations:
\begin{eqnarray*}
    \mathbf{w}_{h_{1},1}, \dots , \mathbf{w}_{h_{m},t}, \mathbf{w}_{c_{1,1},1}, \dots, \mathbf{w}_{c_{n,m},t} = \textsc{Enc}(T), 
\end{eqnarray*} which is then fed into a classification layer with a Softmax activation to assign a score for each token to a class $y \in \mathcal{Y}$. 

\subsection{Data Augmentation} \label{sec:DA}
As mentioned above, existing DA techniques for NER, such as those presented in \cite{dataAugment1}, are not a good fit for tabular data, since they produce augmented tables with inconsistent context.
For example, the common label-wise token replacement (LWTR) may replace the \textit{QUANT} token \textit{nominal} in Figure~\ref{fig:example_table} with \textit{height} or the \textit{UoM} \textit{bar} with \textit{Celsius}. This clearly introduces inconsistencies in the context, since \textit{height pressure} has no physical meaning and neither \textit{height} nor \textit{pressure} are measured in \textit{Celsius}.
A visualization of such an inconsistent table can be seen in the Appendix in Figure~\ref{fig:lwtr}.

\begin{figure*}
    \centering
    \includegraphics[width=0.99\textwidth]{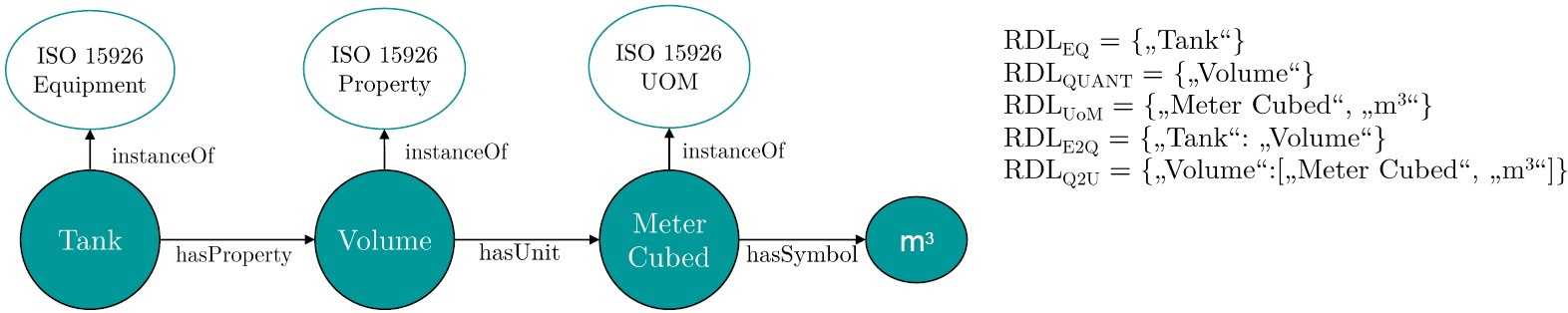}
    \caption{Example graph from POSC Caesar and resulting sets/dictionaries for RDLTab.}
    \label{fig:posc_caesar}
\end{figure*}

To overcome this problem external domain-specific knowledge is needed. For many industrial domains there exist resources (standardized vocabularies, data models) that can be incorporated for DA.
We propose a novel DA approach which leverages existing industrial semantic data models to augment and to generate tables with consistent context. In particular, we use the Reference Data Library (RDL)  
of POSC Caesar (ISO-15926)\footnote{\url{http://data.posccaesar.org/rdl/}}. The RDL is a rich source of a domain-specific vocabulary and relations in the process industry. For example, it defines taxonomies that represent specific types of equipment (\textit{EQ}), but also physical quantities (\textit{QUANT}) that plant equipment typically possess. Figure~\ref{fig:posc_caesar} shows a small excerpt of the RDL as knowledge graph. We leverage this data in the process of augmenting existing tables with consistent equipment, quantity and unit of measure (\textit{UoM}) context as follows.

First, we extract surface names (\textit{sfn}) of all entities of type $\textit{ENT}$ into a respective set $RDL_{ENT} = \left\{\textit{sfn-ent}_1, \textit{sfn-ent}_2, \dots\right\}$, where $\textit{ENT} \in \{\textit{EQ}, \textit{QUANT}, \textit{UoM}\}$. Additionally, we extract a dictionary $RDL_{E2Q}: RDL_{\textit{EQ}} \rightarrow RDL_{\textit{QUANT}}$ that holds a set of applicable quantities for every equipment and a second dictionary $RDL_{Q2U}: RDL_{\textit{QUANT}} \rightarrow RDL_{\textit{UoM}}$ that stores all applicable units of measure for a certain quantity. 
The extracted sets for the example graph are also shown in Figure \ref{fig:posc_caesar}. 

To ease notation we define a function $f_{ner}$ which returns the set of entity types contained in the set of cells passed as arguments, e.g., $f_{ner}(T_{[:,2]}) = \{\textit{EQ}\}$ means that the second column of table $T$ contains entities of type \textit{EQ}.
\paragraph{Augmentation procedure}
Given a table $T$ we generate an augmented sample $T_{aug}$ as follows:
\begin{enumerate}
    \item Sample $k$ columns that contain no NER annotations as starting point for augmentation, $T_{aug} \leftarrow \textit{sample}(\bigcup_j T_{[:,j]}, k)$, where $f_{ner}(T_{[:,j]}) = \varnothing$. 
    
    \item For every row $i$ in $T_{aug}$: An \textit{EQ} entity surface name $\textit{sfn-eq}_i$ is sampled uniformly at random from $RDL_{\textit{EQ}}$. The cells in column $k+1$ hold the sampled names: $c_{i,k+1} \leftarrow \textit{sfn-eq}_i$. 
    
    \item Sample a column header $h_{eq}$ from all training table columns that contain at least one \textit{EQ} annotation: $h_{k+1} \leftarrow h_{eq}$.
    
    \item For every sampled equipment $\textit{sfn-eq}_i$: a \textit{QUANT} entity surface name $\textit{sfn-quant}_i$ is sampled uniformly at random from $RDL_{E2Q}(eq_i)$. Each $\textit{sfn-quant}_i$ is a new column header in $H_{aug} \leftarrow H_{aug} \cup \{\textit{sfn-quant}_i\}$. Fill the resp. cells $c_{i,k+i+1}$ with a random numeric value and optionally a randomly sampled \textit{UoM} surface name from $RDL_{Q2U}(\textit{quant}_i)$.
    
    \item Finally, generate a last column, where for every sampled equipment $\textit{sfn-eq}_i$ an artificial \textit{TAG} entity surface name $\textit{sfn-tag}_i$ is generated. This column's header is then sampled from all training tables headers that contain at least one \textit{TAG} annotation.
\end{enumerate}

Artificial tags are generated by forming an acronym from the \textit{EQ} entity name and adding groups of random alphanumeric strings, optionally divided by the dash '-' character (which is similar to tagging standards).




\section{Experiments}
In this section we empirically study the performance of \textsc{TabNER} compared against several baselines as well as the benefits of our domain-specific table DA technique.

\subsection{Dataset}
We extract 79 tables from a pool of real-world spreadsheets describing industrial plant equipment. To get expert labels, we sub-sampled each table to have a maximum of 5 rows. The labels were collected on a cell-by-cell basis using the tool Prodigy\footnote{\url{https://prodi.gy}}. The statistics of the dataset are shown in Table~\ref{tab:tab:experiments}; the mean number of NER-tags per table is 18, the other columns show the absolute number of NER-tags for each entity type.
All experiments are carried out in a 5-fold cross validation where we use 10\% of each fold's training data as validation~set. 

\begin{table}[t]
\begin{tabular}{ {c} {c} {c} {c} {c} {c} {c} {c}}

		tables & $\mu_{ner}$ &  \textit{TAG} & \textit{EQ} & \textit{QUANT} & \textit{UoM}  \\ \hline
		79 & 18  & 295 & 359 & 427 & 359
\end{tabular}
\caption{Dataset used for experiments.}
\label{tab:tab:experiments}
\end{table}



\subsection{Baselines}
We compare the performance of \textsc{TabNER} to multiple baselines.
First, we design a rule-based NER (\textsc{RuleNER}) based on spaCy's EntityRuler\footnote{\url{https://spacy.io/api/entityruler}} using the same domain-specific vocabularies from RDL as described in section~\ref{sec:DA} for matching. For detecting entities of type \textit{TAG} we employ a heuristic: find the column with most unique body values which does not contain any known vocabulary terms. Then we mark all alphanumeric tokens as \textit{TAG}. 
The second baseline is a \textsc{BiLSTM}-CRF model that uses word embeddings (pre-trained GloVe-6B-100d) as well as character embeddings \cite{ma-hovy-2016}. Here, we simply feed each table in linearized form as input. 
Lastly, we fine-tune a vanilla sequential $\textsc{BERT}$, again with linearized input tables, without any table structure encoding to study if the tabular structure inductive bias is justified.

\paragraph{DA techniques}
We refer to the DA method explained in section~\ref{sec:DA} as RDLTab and compare its performance against LWTR.
For both DA techniques, we experimented with $n=1,2$ number of augmented tables per original table in each epoch. In the case of LWTR, we generate $n$ new tables by randomly replacing $m/2$ tokens, where $m$ is the total number of NER labels available for the table. When applying RDLTab, we generate $n$ new tables for every table in the training set. The best performance was achieved with $n=1$ sample of augmented tables. Therefore, the presented results are with $n=1$ for both DA techniques and the comparison with the performance when $n=2$ samples is discussed in the Appendix.



\section{Results and Analysis}
\begin{figure}
    \includegraphics[width=\linewidth,height=5cm]{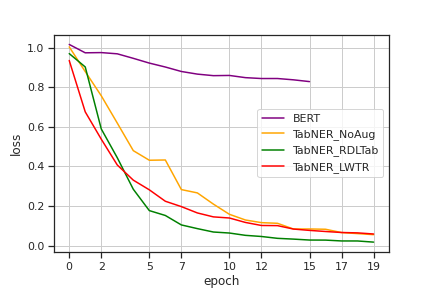}
    \caption{Convergence of the training loss.}
    \label{fig:loss}
\end{figure}

\paragraph{Convergence}
First, we analyze the progress of the training loss to study the convergence of the different NER models, see Figure~\ref{fig:loss}. The loss of vanilla BERT is quite flat from the beginning and after a few epochs gets stuck at a bad local optimum - hits early stopping based on validation. We argue that the global attention and position encoding across the full table are blurring the NER training signal for BERT and since we could not find a setting to make it converge properly, we excluded it from further experiments. A more detailed analysis can be found in the Appendix. 

In contrast to BERT, the training loss for \textsc{TabNER} is converging quickly. Using only the training data, without augmentation, has the least steepest decline, which is due to observing less labels per epoch. LWTR shows a very steep decline in the beginning which, however, flattens out sooner than RDLTab. Our hypothesis here is that LWTR adds helpful variance in the labels at the beginning, but has less variance to add in the long run, since it can only sample from known training tables. RDLTab on the other hand produces a more novel table context over time as the RDL has richer external vocabulary.

\paragraph{Table structure vs. sequential inductive bias} We present the final cross-validation F1 scores in Table~\ref{tab:f1_scores}. It can be seen that $\textsc{TabNER}$ outperforms the baselines in all DA settings, proving the benefits of being biased towards tabular structure. Surprisingly, \textsc{BiLSTM-CRF} does not suffer from the linearized global table context as much as BERT does and still shows competitive performance. One reason might be that the sequential attention in the \textsc{BiLSTM} is trained from scratch and can therefore learn to only focus on very narrow context. While \textsc{BERT} is already pre-trained to take long-range context into account.

\paragraph{Data Augmentation}
The RDLTab DA boosts performance for both \textsc{TabNER} and \textsc{BiLSTM-CRF}. This shows the added value of rich external vocabulary for industrial low-resource problems. Interestingly, LWTR harms performance in both cases. We attribute this to the problem of producing phrases that are non-meaningful physically and inconsistent in a tabular context. 

\begin{table}
    \centering
    \begin{tabular}{l|c|c|c}
    \hline
    \textbf{Model} & \textbf{NoAug} & \textbf{LWTR} & 
    \textbf{RDLTab} \\
    \hline
    \textsc{RuleNER} & 0.08 & - & - \\
    \textsc{BiLSTM-CRF} & 0.53 & 0.46 & 0.55 \\
    \textsc{TabNER} & \textbf{0.54} & \textbf{0.52} & \textbf{0.58} \\
    \hline
    \end{tabular}
    \caption{F1 scores with different DA techniques.}
    \label{tab:f1_scores}
\end{table}

\section{Conclusion}
In this paper, we demonstrate the applicability of TaLMs to a novel NER problem in industrial spreadsheets. Our experiments show that the tabular inductive bias of TaLMs is not only beneficial for this problem, but may even a necessary condition when relying on pre-trained transformer-based models. 
In addition to that we present a DA technique leveraging publicly-available industrial standard information models to produce augmented tables with physically sound and consistent context. Compared to an off-the-shelve DA, this technique shows improved NER performance.

Future work includes understanding how much tabular context is needed to make training large-scale model more efficient. Another fruitful area is active learning for tasks using TaLMs to reduce the time for collecting expert labels. 

\bibliographystyle{acl_natbib}
\bibliography{emnlp2022}

\begin{thebibliography}{21}
\expandafter\ifx\csname natexlab\endcsname\relax\def\natexlab#1{#1}\fi

\bibitem[{Cutrona et~al.(2020)Cutrona, Bianchi, Jiménez-Ruiz, and
  Palmonari}]{cutrona_vincenzo_2020_4246370}
Vincenzo Cutrona, Federico Bianchi, Ernesto Jiménez-Ruiz, and Matteo
  Palmonari. 2020.
\newblock \href {https://doi.org/10.5281/zenodo.4246370} {{Tough Tables:
  Carefully Evaluating Entity Linking for Tabular Data}}.

\bibitem[{Dai and Adel(2020)}]{dataAugment1}
Xiang Dai and Heike Adel. 2020.
\newblock \href {https://doi.org/10.18653/v1/2020.coling-main.343} {An analysis
  of simple data augmentation for named entity recognition}.
\newblock In \emph{Proceedings of the 28th International Conference on
  Computational Linguistics, {COLING} 2020, Barcelona, Spain (Online), December
  8-13, 2020}, pages 3861--3867. International Committee on Computational
  Linguistics.

\bibitem[{Deng et~al.(2020)Deng, Sun, Lees, Wu, and Yu}]{turl}
Xiang Deng, Huan Sun, Alyssa Lees, You Wu, and Cong Yu. 2020.
\newblock Turl: Table understanding through representation learning.
\newblock \emph{Proc. VLDB Endow.}, 14(3):307–319.

\bibitem[{Dong et~al.(2022)Dong, Cheng, He, Zhou, Zhou, Zhou, Liu, Han, and
  Zhang}]{Dong2022}
Haoyu Dong, Zhoujun Cheng, Xinyi He, Mengyu Zhou, Anda Zhou, Fan Zhou, Ao~Liu,
  Shi Han, and Dongmei Zhang. 2022.
\newblock Table pre-training: A survey on model architectures, pre-training
  objectives, and downstream tasks.
\newblock In \emph{IJCAI'2022 SURVEY TRACK}.

\bibitem[{Feng et~al.(2021)Feng, Gangal, Wei, Chandar, Vosoughi, Mitamura, and
  Hovy}]{surveyDA}
Steven~Y. Feng, Varun Gangal, Jason Wei, Sarath Chandar, Soroush Vosoughi,
  Teruko Mitamura, and Eduard~H. Hovy. 2021.
\newblock \href {https://doi.org/10.18653/v1/2021.findings-acl.84} {A survey of
  data augmentation approaches for {NLP}}.
\newblock In \emph{Findings of the Association for Computational Linguistics:
  {ACL/IJCNLP} 2021, Online Event, August 1-6, 2021}, volume {ACL/IJCNLP} 2021
  of \emph{Findings of {ACL}}, pages 968--988. Association for Computational
  Linguistics.

\bibitem[{Herzig et~al.(2020)Herzig, Nowak, M{\"{u}}ller, Piccinno, and
  Eisenschlos}]{tapas}
Jonathan Herzig, Pawel~Krzysztof Nowak, Thomas M{\"{u}}ller, Francesco
  Piccinno, and Julian~Martin Eisenschlos. 2020.
\newblock \href {https://doi.org/10.18653/v1/2020.acl-main.398} {Tapas: Weakly
  supervised table parsing via pre-training}.
\newblock In \emph{Proceedings of the 58th Annual Meeting of the Association
  for Computational Linguistics, {ACL} 2020, Online, July 5-10, 2020}, pages
  4320--4333. Association for Computational Linguistics.

\bibitem[{Ibrahim et~al.(2016)Ibrahim, Riedewald, and Weikum}]{Ibrahim2016}
Yusra Ibrahim, Mirek Riedewald, and Gerhard Weikum. 2016.
\newblock \href {https://doi.org/10.1145/2983323.2983772} {Making sense of
  entities and quantities in web tables}.
\newblock In \emph{Proceedings of the 25th ACM International on Conference on
  Information and Knowledge Management}, CIKM '16, page 1703–1712.

\bibitem[{Koleva et~al.(2021)Koleva, Ringsquandl, Joblin, and
  Tresp}]{Koleva2021}
Aneta Koleva, Martin Ringsquandl, Mitchell Joblin, and Volker Tresp. 2021.
\newblock Generating table vector representations.
\newblock In \emph{CEUR Workshop Proceedings - Deep Learning for Knowledge
  Graphs (DL4KG)}.

\bibitem[{Longpre et~al.(2020)Longpre, Wang, and DuBois}]{Longpre2020}
Shayne Longpre, Yu~Wang, and Chris DuBois. 2020.
\newblock \href {https://doi.org/10.18653/v1/2020.findings-emnlp.394} {How
  effective is task-agnostic data augmentation for pretrained transformers?}
\newblock In \emph{Findings of the Association for Computational Linguistics:
  {EMNLP} 2020, Online Event, 16-20 November 2020}, volume {EMNLP} 2020 of
  \emph{Findings of {ACL}}, pages 4401--4411. Association for Computational
  Linguistics.

\bibitem[{Ma and Hovy(2016)}]{ma-hovy-2016}
Xuezhe Ma and Eduard Hovy. 2016.
\newblock \href {https://doi.org/10.18653/v1/P16-1101} {End-to-end sequence
  labeling via bi-directional {LSTM}-{CNN}s-{CRF}}.
\newblock In \emph{Proceedings of the 54th Annual Meeting of the Association
  for Computational Linguistics (Volume 1: Long Papers)}, pages 1064--1074,
  Berlin, Germany. Association for Computational Linguistics.

\bibitem[{Sennrich et~al.(2016)Sennrich, Haddow, and Birch}]{Sennrich2016}
Rico Sennrich, Barry Haddow, and Alexandra Birch. 2016.
\newblock \href {https://doi.org/10.18653/v1/p16-1009} {Improving neural
  machine translation models with monolingual data}.
\newblock In \emph{Proceedings of the 54th Annual Meeting of the Association
  for Computational Linguistics, {ACL} 2016, August 7-12, 2016, Berlin,
  Germany, Volume 1: Long Papers}. The Association for Computer Linguistics.

\bibitem[{Simard et~al.(1996)Simard, LeCun, Denker, and Victorri}]{Simard1996}
Patrice~Y. Simard, Yann LeCun, John~S. Denker, and Bernard Victorri. 1996.
\newblock \href {https://doi.org/10.1007/3-540-49430-8\_13} {Transformation
  invariance in pattern recognition-tangent distance and tangent propagation}.
\newblock In Genevieve~B. Orr and Klaus{-}Robert M{\"{u}}ller, editors,
  \emph{Neural Networks: Tricks of the Trade}, volume 1524 of \emph{Lecture
  Notes in Computer Science}, pages 239--27. Springer.

\bibitem[{Vaswani et~al.(2017)Vaswani, Shazeer, Parmar, Uszkoreit, Jones,
  Gomez, Kaiser, and Polosukhin}]{Vaswani2017}
Ashish Vaswani, Noam Shazeer, Niki Parmar, Jakob Uszkoreit, Llion Jones,
  Aidan~N. Gomez, Lukasz Kaiser, and Illia Polosukhin. 2017.
\newblock \href
  {https://proceedings.neurips.cc/paper/2017/hash/3f5ee243547dee91fbd053c1c4a845aa-Abstract.html}
  {Attention is all you need}.
\newblock In \emph{Advances in Neural Information Processing Systems 30: Annual
  Conference on Neural Information Processing Systems 2017, December 4-9, 2017,
  Long Beach, CA, {USA}}, pages 5998--6008.

\bibitem[{Wang et~al.(2021)Wang, Dong, Jia, Li, Fu, Han, and Zhang}]{tuta}
Zhiruo Wang, Haoyu Dong, Ran Jia, Jia Li, Zhiyi Fu, Shi Han, and Dongmei Zhang.
  2021.
\newblock \href {https://doi.org/10.1145/3447548.3467434} {{TUTA:} tree-based
  transformers for generally structured table pre-training}.
\newblock In \emph{{KDD} '21: The 27th {ACM} {SIGKDD} Conference on Knowledge
  Discovery and Data Mining, Virtual Event, Singapore, August 14-18, 2021},
  pages 1780--1790. {ACM}.

\bibitem[{Wei and Zou(2019)}]{Wei2019}
Jason~W. Wei and Kai Zou. 2019.
\newblock \href {https://doi.org/10.18653/v1/D19-1670} {{EDA:} easy data
  augmentation techniques for boosting performance on text classification
  tasks}.
\newblock In \emph{Proceedings of the 2019 Conference on Empirical Methods in
  Natural Language Processing and the 9th International Joint Conference on
  Natural Language Processing, {EMNLP-IJCNLP} 2019, Hong Kong, China, November
  3-7, 2019}, pages 6381--6387. Association for Computational Linguistics.

\bibitem[{Wu et~al.(2018)Wu, Hsiao, Cheng, Hancock, Rekatsinas, Levis, and
  R\'{e}}]{Wu2018}
Sen Wu, Luke Hsiao, Xiao Cheng, Braden Hancock, Theodoros Rekatsinas, Philip
  Levis, and Christopher R\'{e}. 2018.
\newblock \href {https://doi.org/10.1145/3183713.3183729} {Fonduer: Knowledge
  base construction from richly formatted data}.
\newblock In \emph{Proceedings of the 2018 International Conference on
  Management of Data}, SIGMOD '18, page 1301–1316, New York, NY, USA.
  Association for Computing Machinery.

\bibitem[{Yaseen and Langer(2021)}]{dataAugment2}
Usama Yaseen and Stefan Langer. 2021.
\newblock \href {http://arxiv.org/abs/2108.11703} {Data augmentation for
  low-resource named entity recognition using backtranslation}.
\newblock \emph{CoRR}, abs/2108.11703.

\bibitem[{Yin et~al.(2020)Yin, Neubig, Yih, and Riedel}]{TaBERT}
Pengcheng Yin, Graham Neubig, Wen{-}tau Yih, and Sebastian Riedel. 2020.
\newblock Tabert: Pretraining for joint understanding of textual and tabular
  data.
\newblock In \emph{Proceedings of the 58th Annual Meeting of the Association
  for Computational Linguistics, {ACL} 2020, Online, July 5-10, 2020}, pages
  8413--8426. Association for Computational Linguistics.

\bibitem[{Yun et~al.(2019)Yun, Han, Chun, Oh, Yoo, and Choe}]{Yun2019}
Sangdoo Yun, Dongyoon Han, Sanghyuk Chun, Seong~Joon Oh, Youngjoon Yoo, and
  Junsuk Choe. 2019.
\newblock \href {https://doi.org/10.1109/ICCV.2019.00612} {Cutmix:
  Regularization strategy to train strong classifiers with localizable
  features}.
\newblock In \emph{2019 {IEEE/CVF} International Conference on Computer Vision,
  {ICCV} 2019, Seoul, Korea (South), October 27 - November 2, 2019}, pages
  6022--6031. {IEEE}.

\bibitem[{Zhang et~al.(2015)Zhang, Zhao, and LeCun}]{Zhang2015}
Xiang Zhang, Junbo~Jake Zhao, and Yann LeCun. 2015.
\newblock \href
  {https://proceedings.neurips.cc/paper/2015/hash/250cf8b51c773f3f8dc8b4be867a9a02-Abstract.html}
  {Character-level convolutional networks for text classification}.
\newblock In \emph{Advances in Neural Information Processing Systems 28: Annual
  Conference on Neural Information Processing Systems 2015, December 7-12,
  2015, Montreal, Quebec, Canada}, pages 649--657.

\bibitem[{Zhong et~al.(2020)Zhong, Lewis, Wang, and Zettlemoyer}]{Zhong2020}
Victor Zhong, Mike Lewis, Sida~I. Wang, and Luke Zettlemoyer. 2020.
\newblock \href {https://doi.org/10.18653/v1/2020.emnlp-main.558} {Grounded
  adaptation for zero-shot executable semantic parsing}.
\newblock In \emph{Proceedings of the 2020 Conference on Empirical Methods in
  Natural Language Processing, {EMNLP} 2020, Online, November 16-20, 2020},
  pages 6869--6882. Association for Computational Linguistics.

\end{thebibliography}

\appendix

\section{Appendix}
\label{sec:appendix}
\begin{figure*}[t]
    \centering
    \includegraphics[width=0.9\textwidth]{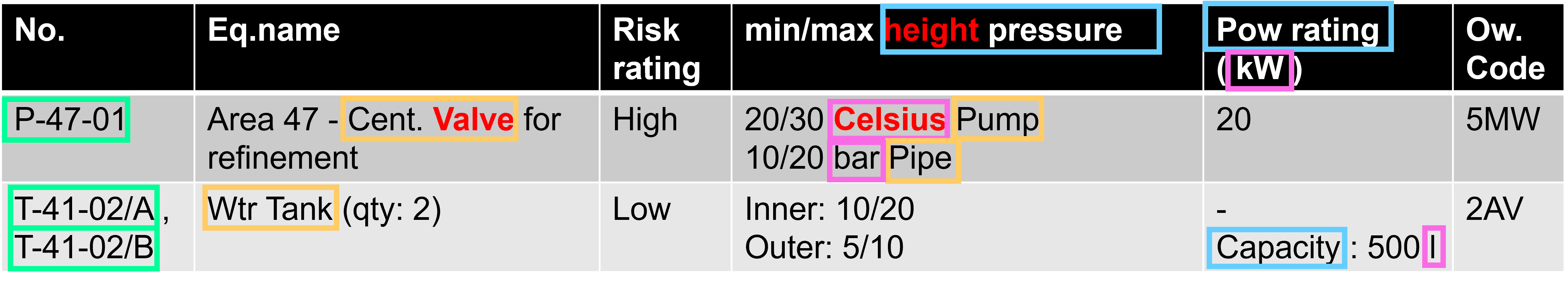}
    \caption{LWTR introduces inconsistent tabular context. Red tokens have been replaced in the original in Figure~\ref{fig:example_table}.}
    \label{fig:lwtr}
\end{figure*}

\paragraph{DA example}
Figure~\ref{fig:lwtr} shows how tabular context becomes inconsistent when applying LWTR to the table in Figure~\ref{fig:example_table}. The red tokens have been replaced with sampled tokens from the training set.
It can be seen that the \textit{QUANT} entity \textit{height pressure} is now physically meaningless and neither \textit{height} nor \textit{pressure} are measured in \textit{Celsius}.

\paragraph{Probing table context}
\begin{figure*}[b]
    \centering
\begin{subfigure}[b]{0.48\textwidth}
    \centering
     \includegraphics[trim=30 0 0 0,width=0.8\textwidth]{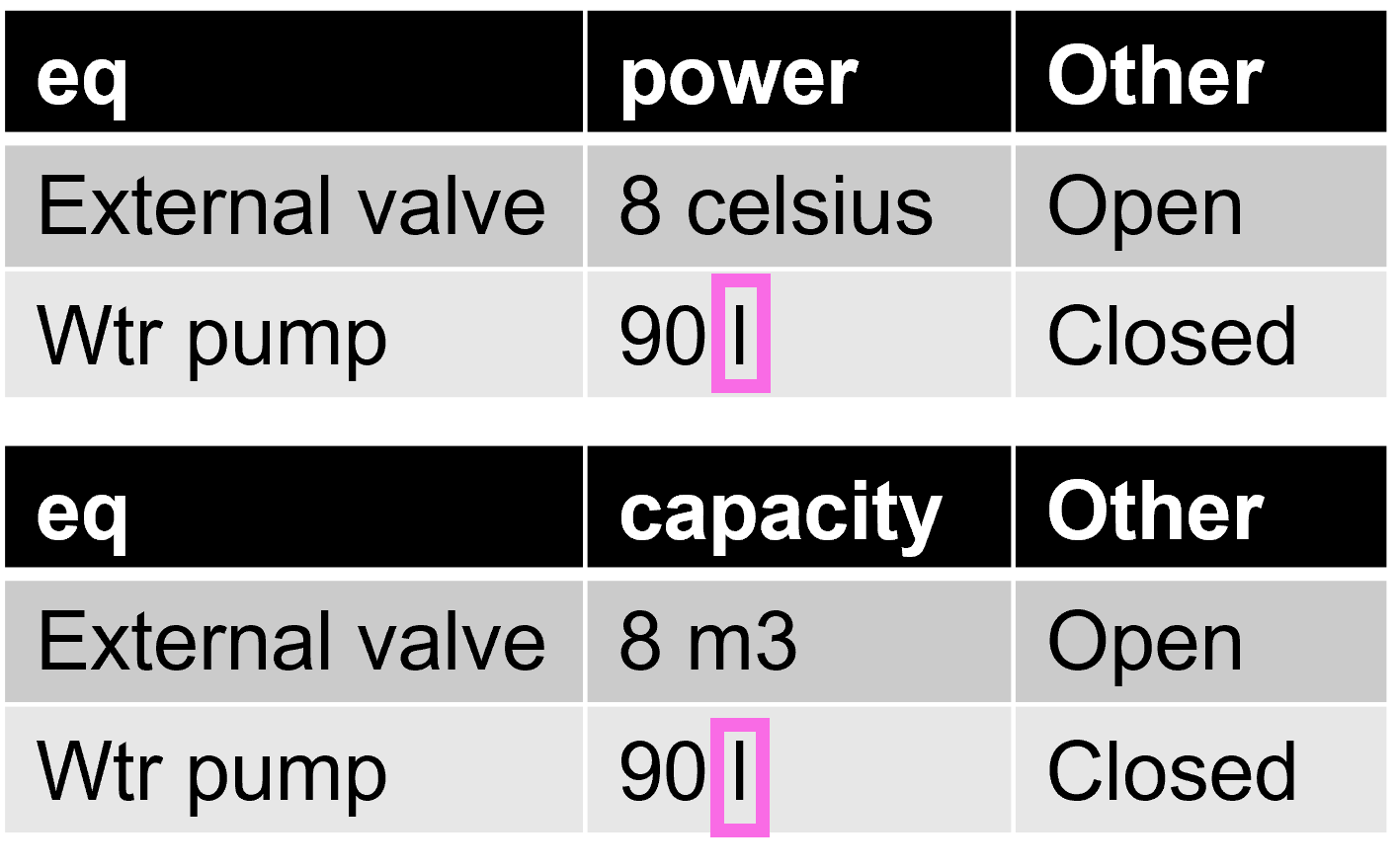}
    \caption{Two synthetic tables with small modifications. Top has random context, bottom has consistent context.}
    \label{fig:appendix_table}
\end{subfigure}
\hfill
\begin{subfigure}[b]{0.48\textwidth}
    \centering
    \includegraphics[width=0.8\textwidth]{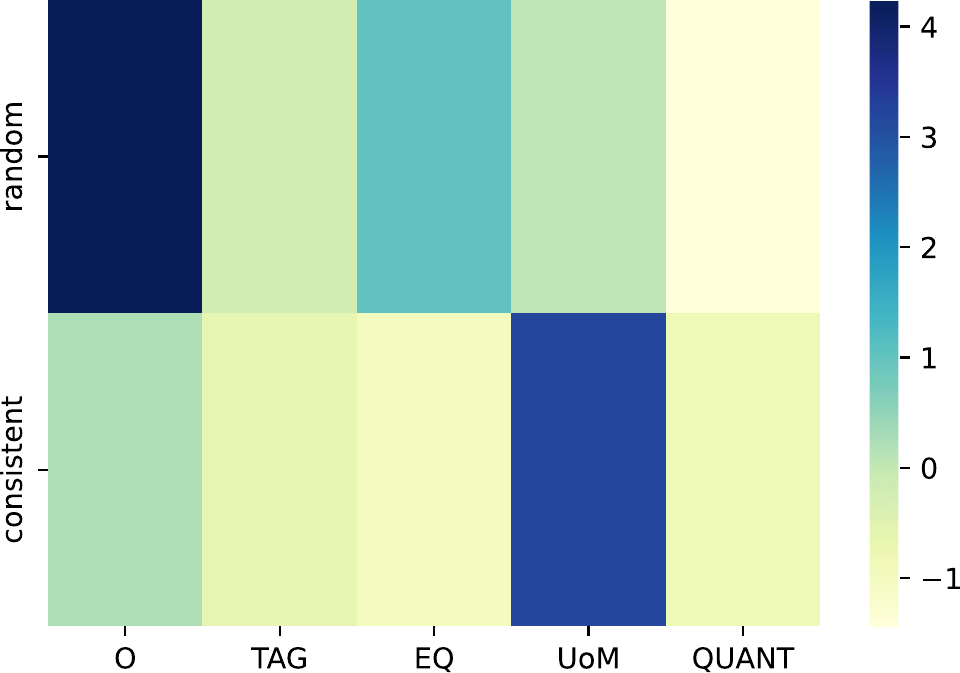}
    \caption{Unnormalized logits for token 'l' in top and bottom table in \ref{fig:appendix_table}.}
    \label{fig:logits}
 \end{subfigure}
 \caption{\textsc{TabNER} token logits with synthetic consistent and randomized  table context.}
\end{figure*}

To demonstrate the sensitivity of \textsc{TabNER} towards table context, we construct two synthetic tables with slightly modified cell content. The table at the top in Figure~\ref{fig:appendix_table} has a column with header \textit{power} (\textit{QUANT}) with body cells having random (inconsistent) \textit{UoM} entities \textit{8 celsius} and \textit{90 l}. The bottom table's column with header \textit{capacity} has consistent \textit{UoM} context. We are interested in how these two different contexts affect the classification of token \textit{'l'}, which is hard to classify without context. In a column like \textit{capacity} it likely refers to the \textit{UoM} entity \textit{'liter'}. However, in most other contexts \textit{'l'} is not part of any entity. Looking at the respective logits in Figure~\ref{fig:logits}, we can see that \textsc{TabNER} is sensitive to these context changes. The highest scoring class for the random context is \textit{O}, while in the consistent case it is the class \textit{UoM}. This is a beneficial property, since it prevents false positives for highly ambiguous tokens such as \textit{'l'}, which only in very specific contexts are likely to be entities.

\paragraph{Experiment Details}
For fair comparison, both \textsc{TabNER} and \textsc{BERT} are based on the pre-trained 'bert-base-uncased' and we select the best hyperparameters from these ranges: learning rate $\{5e^{-5}, 1e^{-5}, 5e^{-4}\}$, batch size $\{2, 4, 8\}$. The learning follows a linearly decreasing schedule with a maximum of 20 epochs. 
For the \textsc{BiLSTM-CRF} we use the NER hyperparmeters from \cite{ma-hovy-2016}.

\begin{figure}[p]
    \includegraphics[width=\linewidth]{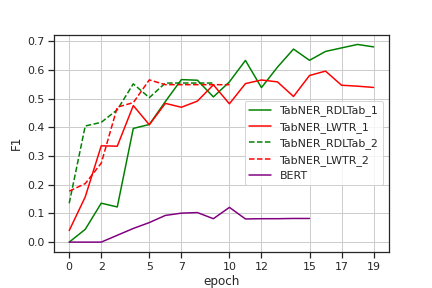}
    \caption{F1 score on validation set during training.}
    \label{fig:f1score}
\end{figure}

\paragraph{BERT Analysis}
In our experiments, we observe that BERT almost exclusively fits to the \textit{O} token labels in the training set and does not pick up on the other NER signals. Since it is an imbalanced problem, our hypothesis is that the global attention and position encoding across the full table blurs tokens with less frequent NER signals and BERT cannot properly fit them. More sophisticated weighted loss functions could be tried to remedy this problem. In Figure~\ref{fig:f1score} the progress of the validation set F1 score is shown. Even though the training loss is still slightly decreasing, the validation NER performance seems to have already peaked. In all hyperparameter settings (even with much lower learning rate $1e^{-7}$) we could not achieve a test F1 score higher than ~0.03.  

\paragraph{Class-wise F1 scores} 
\begin{table}
\centering
\begin{tabular}{l|c|c|c|c}
\hline
\textbf{Model} & \textit{TAG} & \textit{EQ} & 
\textit{QUANT} & \textit{UoM} \\
\hline
\textsc{RuleNER} & 0.1 & 0.09 & 0.04 & 0.1 \\
\textsc{BiLSTM-CRF} & 0.55 & 0.39 & \textbf{0.54} & 0.67 \\
\textsc{TabNER} &\textbf{ 0.60} & \textbf{0.43} & 0.47 & \textbf{0.77} \\
\hline
\end{tabular}
\caption{Class-wise F1 scores.}
\label{tab:classwise}
\end{table}
As more fine-grained analysis, we present the class-wise F1 scores for each model in Table~\ref{tab:classwise}. We can see that the \textsc{TabNER} is better in extracting entities of types \textit{TAG}, \textit{EQ} and \textit{UoM}, while the \textsc{BiLSTM} model is better at classifying entities of type \textit{QUANT}.

\paragraph{Data Augmentation Samples} We experiment with $n=1,2$ samples to evaluate if increasing the training set by more then 100\% will bring benefit to the TabNER model. Figure \ref{fig:f1score} shows the validation set F1 score for the TabNER model with the two DA techniques, LWTR and RDLTab, and the different $n=1,2$ samples. Consistently, for both techniques, when $n=2$ the model converges much faster, after only $5$ epochs, however the performance of the model is worse compared to when we use $n=1$.

\end{document}